\documentclass[a4paper,twocolumn,11pt,accepted=2017-05-09]{quantumarticle}
\pdfoutput=1
\usepackage[utf8]{inputenc}
\usepackage[english]{babel}
\usepackage[T1]{fontenc}
\usepackage{amsmath}
\usepackage{hyperref}

\usepackage{tikz}
\usepackage{lipsum}
\usepackage{float}
\usepackage{algorithm}
\usepackage{algpseudocode}

\usepackage{amssymb}

\begin{document}

\title{Quantum Recurrent Neural Networks with Encoder-Decoder for Time-Dependent Partial Differential Equations}

\author{Yuan Chen}
\affiliation{Computational and Data Science Program, Middle Tennessee State University, Murfreesboro, 37132, TN, USA}
\email{yc3y@mtmail.mtsu.edu}
\author{Abdul Khaliq}
\affiliation{Computational and Data Science Program, Middle Tennessee State University, Murfreesboro, 37132, TN, USA}
\affiliation{Department of Mathematical Science, Middle Tennessee State University, Murfreesboro, 37132, TN, USA}
\author{Khaled M. Furati}
\affiliation{Department of Mathematics, King Fahd University of Petroleum \& Minerals, Dhahran, 31261, Saudi Arabia}
\maketitle

\begin{abstract}
Nonlinear time-dependent partial differential equations are essential in modeling complex phenomena across diverse fields, yet they pose significant challenges due to their computational complexity, especially in higher dimensions. This study explores Quantum Recurrent Neural Networks within an encoder-decoder framework, integrating Variational Quantum Circuits into Gated Recurrent Units and Long Short-Term Memory networks. Using this architecture, the model efficiently compresses high-dimensional spatiotemporal data into a compact latent space, facilitating more efficient temporal evolution. We evaluate the algorithms on the Hamilton-Jacobi-Bellman equation, Burgers' equation, the Gray-Scott reaction-diffusion system, and the three dimensional Michaelis-Menten reaction-diffusion equation. The results demonstrate the superior performance of the quantum-based algorithms in capturing nonlinear dynamics, handling high-dimensional spaces, and providing stable solutions, highlighting their potential as an innovative tool in solving challenging and complex systems.
\end{abstract}

\section{Introduction}
Partial differential equations (PDEs) are fundamental mathematical tools for modeling diverse phenomena in many fields such as physics, biology, chemistry, and economics. However, for many complex and high-dimensional PDEs, analytical solutions are often unattainable due to their intricate structures and non-linearities. To address this, numerical methods such as the finite-difference method (FDM)~\cite{ozisik2017}, finite-element method (FEM)~\cite{dhatt2012}, and finite-volume method (FVM)~\cite{eymard2000} have been developed to approximate solutions. These techniques have been effective in a variety of applications but face limitations in computational complexity, stability, and scalability, especially when applied to non-linear or high-dimensional problems.

In recent years, deep learning has emerged as a powerful data-driven approach for solving partial differential equations (PDEs), particularly when traditional numerical methods struggle with the complexities of real-world data. Techniques like physics-informed neural networks (PINNs) and their extensions \cite{raissi2019, liu2023, sharma2023} excel by learning complex nonlinear relationships and managing high-dimensional data. By embedding the underlying physical laws directly as constraints within the network architecture, these methods achieve remarkably accurate approximations of PDE solutions.

With the development of neural networks, Recurrent neural networks (RNNs) provided another promising approach in the context of PDEs, especially for problems involving temporal dynamics. RNNs were first introduced by Rumelhart et al. \cite{rumelhart1986} as a class of neural networks designed for processing sequential data by incorporating feedback loops \cite{bahdanau2015, graves2013, sutskever2014}. While standard RNNs demonstrated potential, they suffered from issues like vanishing gradients, which limited their ability to capture long-term dependencies. To address this, long short-term memory (LSTM) networks were introduced by Hochreiter et al. \cite{hochreiter1997} as an extension of RNNs with gating mechanisms that enable effective learning of long-term dependencies. Later, gated recurrent units (GRUs) were proposed by Cho et al.\cite{cho2014} as a simplified alternative to LSTMs, retaining similar capabilities but with fewer parameters.

The application of RNNs and their variants to PDEs has shown significant promise. These models can capture temporal dependencies and approximate complex solutions by learning from data-driven or hybrid approaches. For instance, a physics-incorporated convolutional recurrent neural network for identifying sources and forecasting dynamical systems was proposed by Saha et al \cite{saha2021}. Neural-PDE, an RNN-based framework specifically tailored for solving time-dependent PDEs introduced by Hu et al. \cite{hu2020}. These approaches demonstrate the versatility and potential of RNN-type algorithms in addressing the challenges posed by complex PDE systems. Meanwhile, another tool caught our eyes, an encoder-decoder architectures were proposed by Cho et al. \cite{cho2014encoder}, which could help solving PDEs by compressing high-dimensional spatial-temporal data into a lower-dimensional latent space, enabling the model to efficiently learn complex temporal dynamics and reconstruct accurate high-dimensional solutions. Other RNNs based PDE method can be found in \cite{sirignano2018dgm, wu2022}.

Parallel to the advancements in deep learning, quantum machine learning has emerged as a transformative technology that leverages quantum mechanics for computation. Recent studies have extended these quantum methods to tackle PDEs; Hunout et al. \cite{Hunout2024} introduced a variational quantum algorithm based on Lagrange polynomial encoding, Choi and Ryu \cite{Choi2024} developed a method for multi-dimensional Poisson equations, and Childs et al. \cite{Childs2021} demonstrated high-precision quantum algorithms for PDEs. Moreover, Variational Quantum Circuits (VQCs) have shown significant potential in enhancing classical neural network architectures by incorporating quantum entanglement and superposition \cite{mitarai2018quantum, schuld2018circuit, chen2020variational}. This progress has paved the way for Quantum Recurrent Neural Networks (QRNNs), such as Quantum Long Short-Term Memory (QLSTM) \cite{chen2020quantum} and Quantum Gated Recurrent Units (QGRU) \cite{chen2022reservoir}, which are particularly adept at modeling complex temporal dependencies. Motivated by these advances, this study explores the application of QRNNs to solve challenging non-linear time-dependent PDEs.

This paper is organized as follows: Section 2 presents the methodology, including an overview of both classical and quantum RNN models. Section 3 covers four numerical experiments: the Burgers' equation, Gray-Scott reaction-diffusion system, Hamilton-Jacobi-Bellman (HJB) equation, and 3D Michaelis-Menten reaction-diffusion system. Finally, Section 4 provides the conclusion.

\section{Methodology}

In this work, we utilize a composite model architecture comprising three main components: an encoder, a quantum recurrent neural network (QGRU or QLSTM), and a decoder. The classical auto-encoder compresses higher-dimensional PDE snapshots into a lower-dimensional latent space. These latent representations are then used as inputs to an RNN (in this case, a quantum-classical RNN variant) that learns the temporal evolution of the PDEs. Finally, the RNN’s predictions are decoded back to the original dimensional space, enabling full reconstruction and analysis of the time-dependent PDEs solution. To provide a more comprehensive understanding of our approach, the following sections first introduce the QLSTM and QGRU architectures, followed by a detailed explanation of the encoder-decoder structure.

\subsection{Quantum Long Short-Term Memory (QLSTM)}
\label{subsec1}

\begin{figure}[H]
\includegraphics[width=0.45\textwidth]{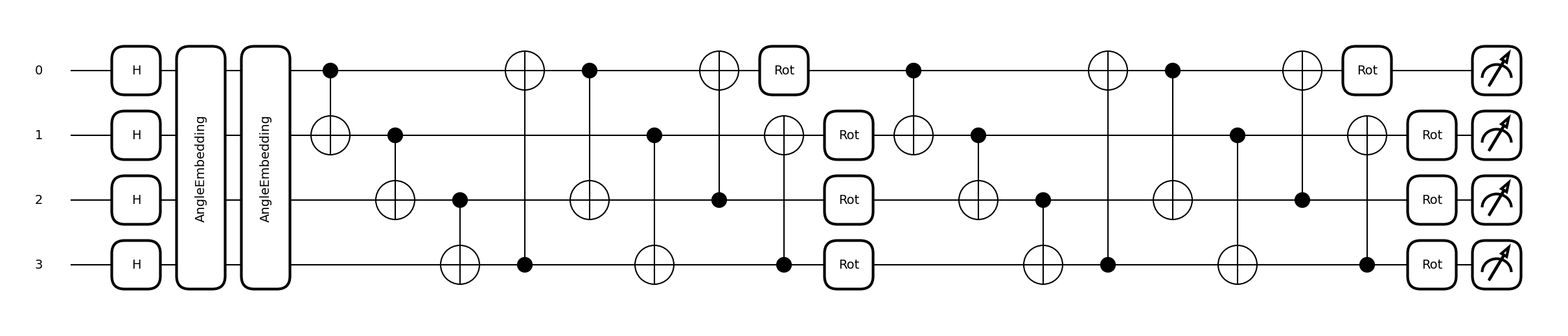}
\caption{An example of VQC architecture with two layers of entanglements for QLSTM.}
\label{fig:vqc}
\end{figure}

VQCs form the core of quantum recurrent neural networks, comprising three primary parts: an encoding layer, a variational layer, and a quantum measurement layer. Figure \ref{fig:vqc} provides an example of a VQC.

On the left of Figure~\ref{fig:vqc}, the encoding layer applies a Hadamard gate and two angle embeddings (treated as $R_y$ and $R_z$ rotations in quantum computing). The central portion is the variational (entanglement) layer. Note that both the number of qubits (4 in this instance) and the number of variational layers (2 here) are adjustable parameters that can be tuned to improve model performance.

The encoding layer is tasked with mapping classical data into quantum amplitudes. This process starts with initializing the quantum circuit in the ground state, then applying Hadamard gates to each qubit to create an unbiased initial state. The state of an $N$-qubit system can be expressed as:
\begin{align}
|\psi\rangle = \sum_{(q_1, q_2, \ldots, q_N) \in \{0, 1\}^N} c_{q_1, q_2, \ldots, q_N} \, |q_1\rangle \otimes |q_2\rangle \otimes \dots \otimes |q_N\rangle,
\end{align}
where $c_{q_1, \ldots, q_N} \in \mathbb{C}$ is the amplitude of each basis state and $\otimes$ denotes the tensor product. By Born's rule, the probability of measuring any specific state is determined by the square of its amplitude’s magnitude:
\begin{align}
\sum_{(q_1, q_2, \ldots, q_N) \in \{0, 1\}^N} \left\lVert c_{q_1, \ldots, q_N} \right\rVert^2 = 1.
\end{align}

Applying the Hadamard gate $H$ on each qubit $|0\rangle$ transforms the circuit into a uniform superposition state. Here, $i$ indexes through all possible $N$-qubit basis states. For each data input $x_i$, the two-angle encoding scheme assigns rotation angles $\arctan(x_i)$ and $\arctan(x_i^2)$, which correspond to $R_y$ and $R_z$ gates respectively. 

\begin{align}
(H|0\rangle)^{\otimes N} = \left(\tfrac{1}{\sqrt{2}}\bigl(|0\rangle + |1\rangle\bigr)\right)^{\otimes N} 
= \tfrac{1}{\sqrt{2^N}} \sum_{i=0}^{2^N-1} |i\rangle.
\end{align}

In this work, the template $\texttt{qml.templates.AngleEmbedding}$ is employed for these rotations, offering flexibility in selecting rotation axes (\emph{e.g.}, $R_x$, $R_y$, or $R_z$).

The variational circuit forms the trainable portion of the VQC and includes parameterized unitary operations. This portion incorporates multiple CNOT gates to produce entanglement, followed by unitary rotations governed by learnable parameters $\alpha, \beta,$ and $\gamma$. In practice, one can repeat this variational block multiple times to enlarge the parameter space and increase expressive power.

\begin{figure}[H]
\centering
\includegraphics[width=0.5\textwidth]{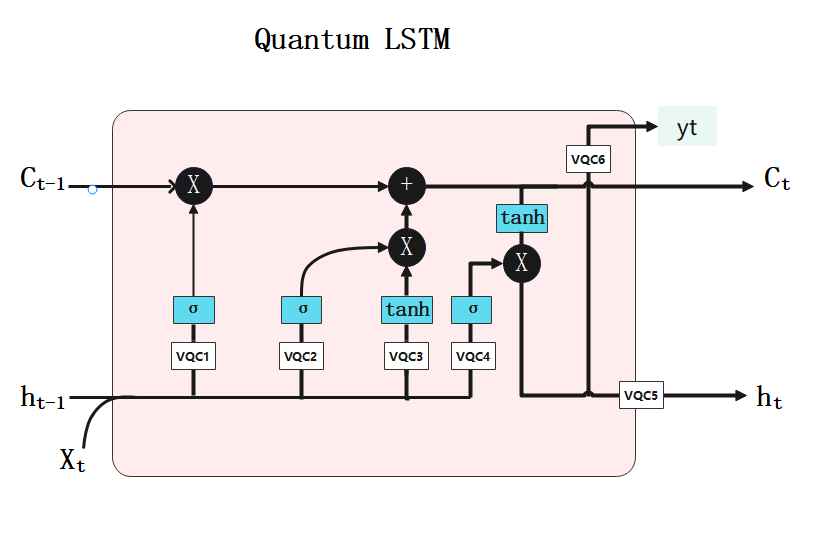}
\caption{Structure of QLSTM.}
\label{fig:qlstm}
\end{figure}

Quantum measurement translates the quantum circuit’s state back into classical information. Owing to quantum systems' inherent probabilistic nature, repeated measurements yield different bit strings. The expectation value of a Hermitian operator $\hat{O}$ with respect to $|\psi\rangle$ is defined as:
\begin{align}
E[\hat{O}] = \langle \psi | \hat{O} | \psi \rangle.
\end{align}
These expectations may be calculated either analytically (in simulation) or derived via multiple shots on actual quantum hardware, incorporating specific noise models. Figure~\ref{fig:qlstm} depicts a single QLSTM with 6 VQCs.

Similar to classical LSTMs, QLSTMs maintain two primary memory elements: the hidden state $h_t$ and the internal cell state $c_t$. The core operations in a QLSTM cell can be written as follows:
\begin{align*}
f_t &= \sigma \bigl(\mathrm{VQC1}(v_t)\bigr),  \\
i_t &= \sigma \bigl(\mathrm{VQC2}(v_t)\bigr),  \\
\tilde{C}_t &= \tanh \bigl(\mathrm{VQC3}(v_t)\bigr), \\
c_t &= f_t \ast c_{t-1} + i_t \ast \tilde{C}_t,  \\
o_t &= \sigma \bigl(\mathrm{VQC4}(v_t)\bigr),  \\
h_t &= \mathrm{VQC5}\bigl(o_t \ast \tanh (c_t)\bigr),  \\
\tilde{y}_t &= \mathrm{VQC6}\bigl(o_t \ast \tanh(c_t)\bigr), \\
y_t &= \mathrm{NN}(\tilde{y}_t),
\end{align*}
where $\ast$ denotes element-wise multiplication. At each timestep, the QLSTM cell takes $v_t$ as input, which is the concatenation of $h_{t-1}$ and $x_t$ (the previous hidden state and current input, respectively).

\subsection{Quantum Gated Recurrent Unit (QGRU)}
\label{subsec2}

\begin{figure}[H]
\hspace{-12pt}\includegraphics[width=0.5\textwidth]{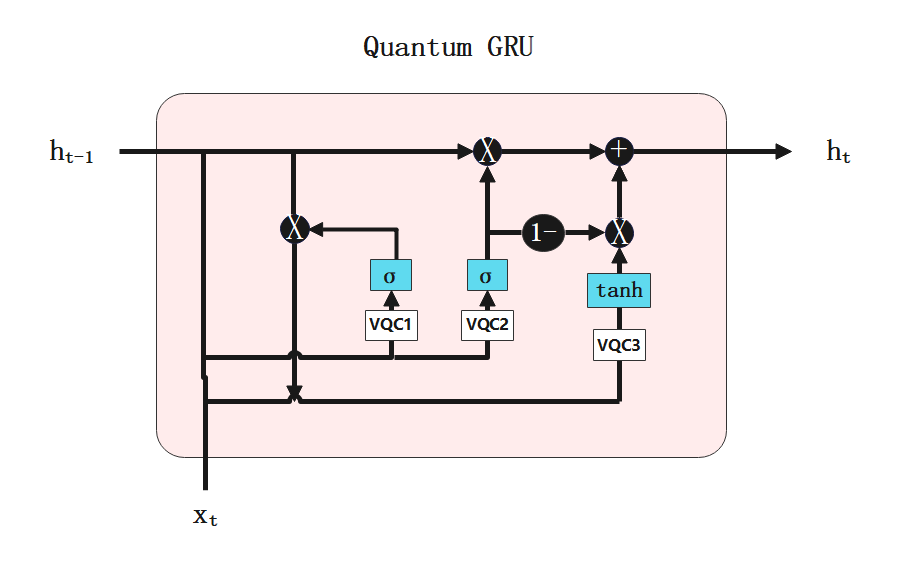}
\caption{Structure of QGRU.}
\label{fig:qgru}
\end{figure}

Similar to QLSTMs, QGRUs enhance the classical GRU architecture by integrating VQCs. The example plot of a QGRU is shown in Figure~\ref{fig:qgru}.

QGRU follows these equations:
\begin{align*}
r_t &= \sigma(\mathrm{VQC1}(v_t)), \\
z_t &= \sigma(\mathrm{VQC2}(v_t)), \\
o_t &= \mathrm{cat}\bigl(x_t,\, r_t \ast H_{t-1}\bigr), \\
\tilde{H}_t &= \tanh\bigl(\mathrm{VQC3}(o_t)\bigr), \\
H_t &= z_t \ast H_{t-1} \;+\; (1 - z_t) \ast \tilde{H}_t, \\
y_t &= \mathrm{NN}(H_t).
\end{align*}

Here, $r_t$ and $z_t$ are the reset and update gates at time step~$t$. The hidden state $H_t$ evolves based on $z_t$ and the candidate hidden state $\tilde{H}_t$, while $v_t$ denotes the concatenation of the previous hidden state $H_{t-1}$ with the current input $x_t$. Because QGRUs employ fewer quantum circuits than QLSTMs, they are typically more computationally efficient. Specifically, three VQCs are used to process quantum information within the gates and candidate hidden state computations. The final output $y_t$ arises from a classical neural network acting on the hidden state $H_t$.

\subsection{Encoder-Decoder Architecture with Quantum Recurrent Models}
\label{subsec3}

The encoder-decoder framework is central to efficiently modeling time-dependent PDEs. The encoder compresses high-dimensional spatial data into a lower-dimensional latent space, enabling the recurrent model to learn temporal patterns efficiently. The encoder maps input data $\mathbf{X} \in \mathbb{R}^{n_{\text{input}}}$ to a latent representation $\mathbf{Z} \in \mathbb{R}^{n_{\text{latent}}}$ using fully connected layers:
\[
\mathbf{Z} = f_{\text{enc}}(\mathbf{X}; \theta_{\text{enc}}),
\]
where $\theta_{\text{enc}}$ represents the encoder parameters.

In the latent space, the temporal dynamics are modeled using QRNNs as we introduced. These quantum-enhanced recurrent models incorporate VQCs to capture complex temporal dependencies and non-linearities. Outputs from the recurrent model are transformed back into the high-dimensional space through the decoder, which reconstructs the original spatial resolution:
\[
\hat{\mathbf{X}}_{t+1} = f_{\text{dec}}(\mathbf{Z}_{t+1}; \theta_{\text{dec}}).
\]
This architecture reduces computational complexity, ensures accurate spatiotemporal predictions, and leverages quantum circuits to enhance expressiveness, making it highly effective for solving time-dependent PDEs.

In our study, we use an Multi-Layer Perceptrons (MLPs) structure for the both encoder and decoder, which is a fundamental neural network architectures consisting of stacked layers of neurons with learnable parameters and activation functions \cite{rumelhart1986}. Fully connected layers are utilized where each neuron is connected to every neuron in the subsequent layer, allowing the network to capture complex hierarchical representations. 

More precisely, our autoencoder structured as deep MLP networks. The encoder progressively reduces the input dimensionality, passing through layers of sizes 4096, 2048, 1024, and 512 before reaching a compressed latent space. Each layer employs the hyperbolic tangent (\(\tanh\)) activation function. After the RNN models are done the prediction works in latent space. The decoder then reconstructs the input by mirroring the encoder structure, expanding the latent representation back to the original input size. A final sigmoid activation ensures the output values remain normalized between 0 and 1. 

Here we show the pseudocode of proposed algorithm.

\begin{algorithm}[H]
\caption{Encoder-decoder QRNNs for Solving Time-Dependent PDEs}
\label{alg:qrnn}
\begin{algorithmic}[1]
\State \textbf{Input:} Discretized PDE data $\mathbf{V}$ with dimensions $(n_t, n_x, n_y)$; 
QRNN parameters $\theta_{\text{qrnn}}$; Encoder parameters $\theta_{\text{enc}}$; Decoder parameters $\theta_{\text{dec}}$.
\State \textbf{Output:} Predicted solutions of PDEs $\hat{\mathbf{V}}$.

\State Normalize the input data $\mathbf{V}$ to obtain $\mathbf{V}_{\text{norm}}$.
\State Flatten the spatial dimensions of $\mathbf{V}_{\text{norm}}$.
\State Encode $\mathbf{V}_{\text{norm}}$ into latent space $\mathbf{Z}$ using the encoder.
\State Generate training sequences $\mathbf{X}_{\text{train}}$ and targets $\mathbf{Y}_{\text{train}}$.
\State Randomly initialize QRNN parameters $\theta_{\text{qrnn}}$.
\State Train QRNN to approximate latent dynamics $\tilde{\mathbf{Z}}$.
\State Decode $\tilde{\mathbf{Z}}$ into high-dimensional predictions $\hat{\mathbf{X}}$ using the decoder.
\State Compute loss $\mathcal{L}$ and minimize it using Adam optimizer.
\State De-normalize predictions $\hat{\mathbf{X}}$ to match the original scale.
\State \textbf{Return:} Final reconstructed predictions $\hat{\mathbf{V}}$.
\end{algorithmic}
\end{algorithm}

\section{Numerical Experiments}
In this section, we investigate the performance of two quantum-enhanced encoder-decoder recurrent neural network architectures, QLSTM and QGRU in solving several nonlinear PDEs. We also evaluated the performance of a classical LSTM for comparison.

\subsection{Experiment 1: Burgers' Equation}
\label{subsec5}

The 2D Burgers' equation \cite{burgers1948} is a nonlinear partial differential equation widely used in fluid dynamics to model viscous flow and turbulence. It is represented as a coupled system:
\begin{align}
    \frac{\partial u}{\partial t} + u \frac{\partial u}{\partial x} + v \frac{\partial u}{\partial y} &= \nu \left( \frac{\partial^2 u}{\partial x^2} + \frac{\partial^2 u}{\partial y^2} \right), \\
    \frac{\partial v}{\partial t} + u \frac{\partial v}{\partial x} + v \frac{\partial v}{\partial y} &= \nu \left( \frac{\partial^2 v}{\partial x^2} + \frac{\partial^2 v}{\partial y^2} \right),
\end{align}

where \( u(x, y, t) \) and \( v(x, y, t) \) represent the velocity fields in the \( x \)- and \( y \)-directions, \( \nu \) is the kinematic viscosity, and \( t \) is the time variable. The spatial domain of the problem consists of a two-dimensional computational grid discretized over \( [0,1] \times [0,1] \) with \( 64 \times 64 \) grid points in the \( x \) and \( y \) directions, respectively.

In experiment 1, we initialize the fields \( u \) and \( v \) with sinusoidal conditions to mimic complex dynamics:
\begin{align}
    u(x, y, 0) &= \sin(2 \pi x) \sin(2 \pi y), \\
    v(x, y, 0) &= \sin(\pi x) \sin(\pi y).
\end{align}
The numerical solution to the Burgers' equation provides the training and test data for the models.

All the models were trained on time-series data generated from the simulation of the 2D Burgers' equation. Key hyperparameters were chosen as follows:
\begin{table}[H]
\centering
\begin{tabular}{|l|l|}
\hline
\textbf{Parameter}            & \textbf{Value} \\ \hline
\textbf{Hidden Size}          & 4              \\ \hline
\textbf{Number of Layers}     & 5              \\ \hline
\textbf{Number of Qubits}     & 4              \\ \hline
\textbf{Number of Quantum Layers} & 4          \\ \hline
\textbf{Learning Rate}        & 0.01           \\ \hline
\textbf{Epochs}               & 50             \\ \hline
\end{tabular}
\caption{Hyperparameters used in the experiments.}
\label{tab:hyperparameters}
\end{table}

All the models are evaluated using Mean Absolute Error (MAE) and Root Mean Square Error (RMSE) to assess performance. The Mean Absolute Error (MAE) and Root Mean Square Error (RMSE) for each model are calculated as follows:
\begin{equation}
    \text{MAE} = \frac{1}{n} \sum_{i=1}^{n} |y_i - \hat{y}_i|,
\end{equation}
\begin{equation}
    \text{RMSE} = \sqrt{\frac{1}{n} \sum_{i=1}^{n} (y_i - \hat{y}_i)^2},
\end{equation}
where \( y_i \) are the true values, \( \hat{y}_i \) are the predictions, and \( n \) is the sample size. MAE focuses on the absolute deviations, while RMSE emphasizes larger errors, making it sensitive to larger deviations.

The following plots illustrate the evolution of the MAE and RMSE metrics for the training and testing phases across epochs.
\begin{figure}[H]
    \centering
    \includegraphics[width=0.45\textwidth]{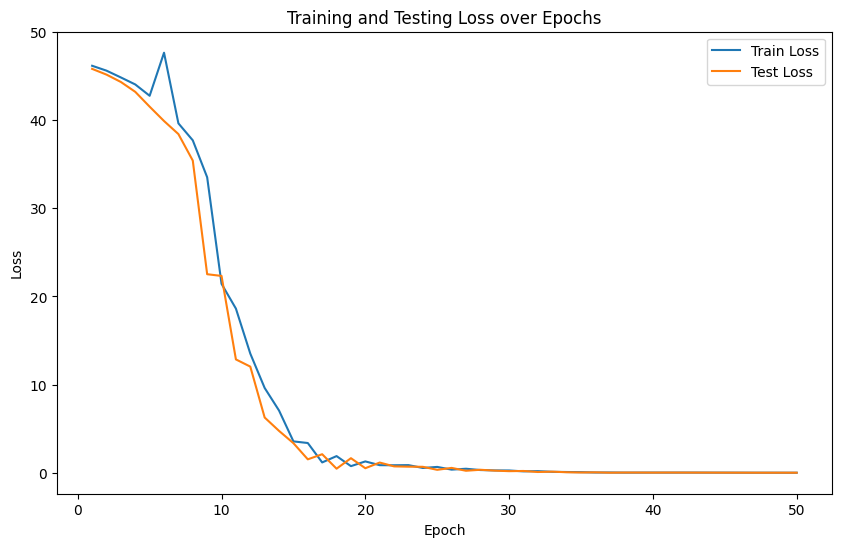}
    \caption{Training and test loss over epochs for LSTM model.}
    \label{fig:log_loss_lstm}
\end{figure}

\begin{figure}[H]
    \centering
    \includegraphics[width=0.45\textwidth]{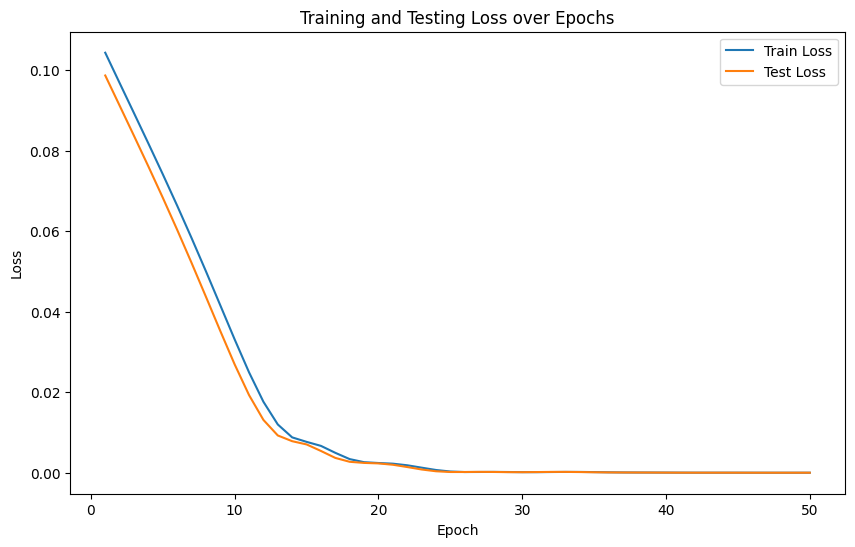}
    \caption{Training and test loss over epochs for QLSTM model.}
    \label{fig:log_loss_qlstm}
\end{figure}

\begin{figure}[H]
    \centering
    \includegraphics[width=0.45\textwidth]{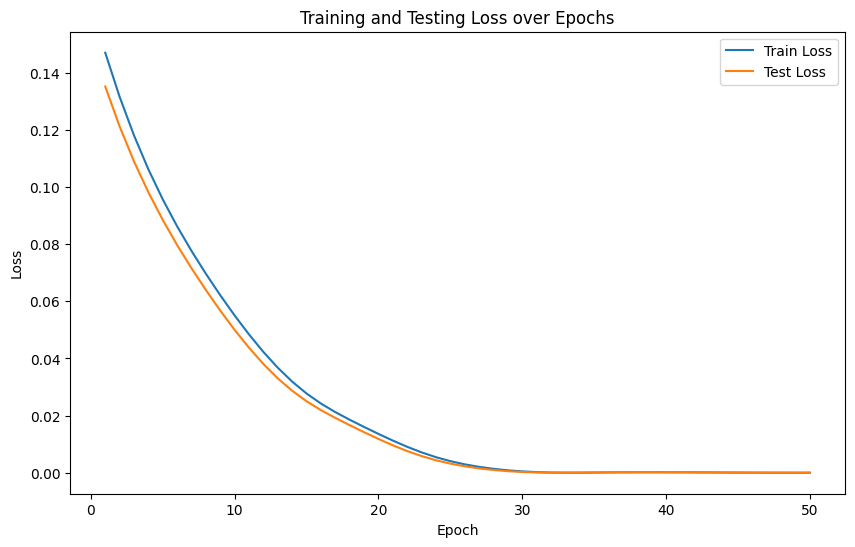}
    \caption{Training and test loss over epochs for QGRU model.}
    \label{fig:log_loss_qgru}
\end{figure}

All models become stable at 30 epochs. After 50 epochs, the final MAE and RMSE for the test sets of all the approaches are summarized in Table~\ref{tab:results_b}.

\begin{table}[H]
\centering
\begin{tabular}{|c|c|c|}
\hline
\textbf{Model} & \textbf{MAE} & \textbf{RMSE} \\ \hline
\textbf{LSTM}  & \(1.582 \times 10^{-3}\) & \(2.137 \times 10^{-3}\) \\ \hline
\textbf{QLSTM} & \(5.316 \times 10^{-4}\) & \(8.110 \times 10^{-4}\) \\ \hline
\textbf{QGRU}  & \(2.981 \times 10^{-4}\) & \(4.364 \times 10^{-4}\) \\ \hline
\end{tabular}
\caption{Experiment 1: Final MAE and RMSE values for the models.}
\label{tab:results_b}
\end{table}

\begin{figure}[H]
    \centering
    \includegraphics[width=0.4\textwidth]{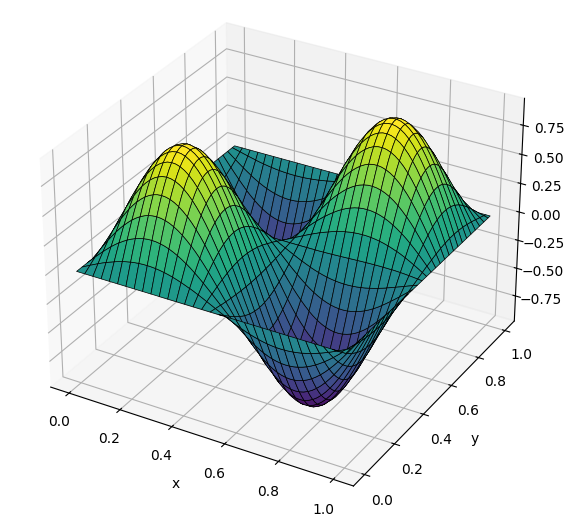}
    \caption{Surface plot of actual data for the 2D Burgers' equation.}
    \label{fig:actual_burger_3d}
\end{figure}

\begin{figure}[H]
    \centering
    \includegraphics[width=0.4\textwidth]{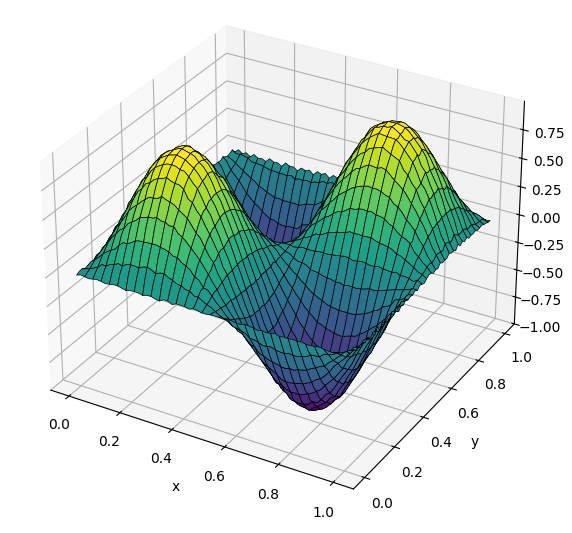}
    \caption{Surface plot of LSTM predictions for the 2D Burgers' equation.}
    \label{fig:lstm_burger_3d}
\end{figure}

\begin{figure}[H]
    \centering
    \includegraphics[width=0.4\textwidth]{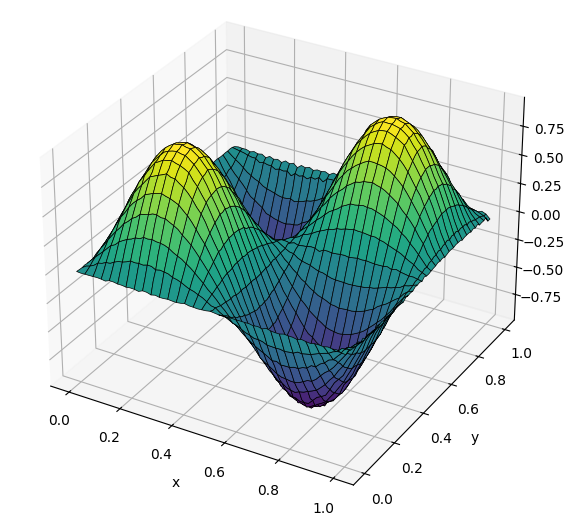}
    \caption{Surface plot of QLSTM predictions for the 2D Burgers' equation.}
    \label{fig:qlstm_burger_3d}
\end{figure}

\begin{figure}[H]
    \centering
    \includegraphics[width=0.4\textwidth]{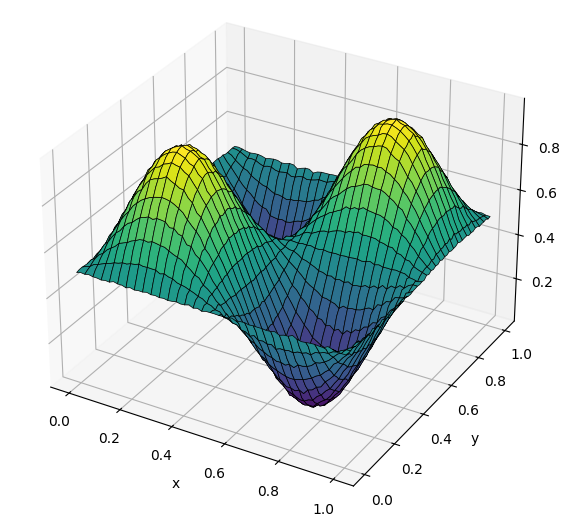}
    \caption{Surface plot of QGRU predictions for the 2D Burgers' equation.}
    \label{fig:qgru_burger_3d}
\end{figure}

The results in Table~\ref{tab:results_b} clearly show that quantum-enhanced recurrent neural networks outperform the classical LSTM in both MAE and RMSE. This demonstrates the superior capability of QRNNs in capturing complex spatiotemporal patterns. Additionally, we present surface plots as shown in Figures \ref{fig:actual_burger_3d} to \ref{fig:qgru_burger_3d}.


The QGRU demonstrates the most accurate and smooth predictions among the models, whereas the QLSTM and LSTM exhibit more fragmented details near the edges. Nonetheless, all models successfully capture the overall trend.
\subsection{Experiment 2: Gray-Scott Reaction-Diffusion System}
\label{subsec6}
The Gray-Scott model \cite{gray1984} is governed by the following partial differential equations, which describe the evolution of two chemical species, \( u \) and \( v \), over time:
\begin{align}
    \frac{\partial u}{\partial t} &= D_u \nabla^2 u - uv^2 + F(1 - u), \\
    \frac{\partial v}{\partial t} &= D_v \nabla^2 v + uv^2 - (F + k)v,
\end{align}

The variables \( u \) and \( v \) denote the concentrations of two interacting chemical species in the system. The diffusion coefficients, \( D_u \) and \( D_v \), characterize the rate at which species \( u \) and \( v \) diffuse through the spatial domain, respectively. The parameter \( F \) represents the feed rate of species \( u \), indicating the influx of this chemical into the system. Additionally, \( k \) denotes the removal rate of species \( v \), determining how quickly it is depleted from the system. Together, these parameters govern the reaction-diffusion dynamics, shaping the spatial and temporal evolution of the species' concentrations.

The system begins with initial random perturbations and evolves into either stable or dynamic patterns based on the values of \( D_u \), \( D_v \), \( F \), and \( k \). The Gray-Scott model, known for its ability to generate self-organizing structures such as stripes, spots, and maze-like patterns, is simulated on a \( 64 \times 64 \) spatial grid. In this experiment, the parameters are set to \( D_u = 0.16 \), \( D_v = 0.08 \), \( F = 0.035 \), \( k = 0.060 \), with spatial and temporal discretization steps of \( dx = 1.0 \) and \( dt = 1.0 \), respectively.

The simulation runs for 1000 time steps, and the concentrations \( u \) and \( v \) are recorded every 10 steps, generating a time series of spatial data. The QRNNs are trained to predict the future evolution of the concentration fields based on a sequence of previous time steps.
The key hyperparameters for the models are the same as experiment 1.

After 100 epochs, the final MAE and RMSE for the predictions are summarized in Table~\ref{tab:results_g1} and Table~\ref{tab:results_g2}.

\begin{table}[H]
\centering
\begin{tabular}{|c|c|c|}
\hline
\textbf{Model} & \textbf{MAE} & \textbf{RMSE} \\ \hline
\textbf{LSTM}  & \(5.256 \times 10^{-5}\) & \(8.669 \times 10^{-5}\) \\ \hline
\textbf{QLSTM} & \(1.509 \times 10^{-5}\) & \(2.130 \times 10^{-5}\) \\ \hline
\textbf{QGRU}  & \(5.747 \times 10^{-5}\) & \(7.821 \times 10^{-5}\) \\ \hline
\end{tabular}
\caption{Experiment 2: Final MAE and RMSE values for Methods, Variable \( u \).}
\label{tab:results_g1}
\end{table}

\begin{table}[H]
\centering
\begin{tabular}{|c|c|c|}
\hline
\textbf{Model} & \textbf{MAE} & \textbf{RMSE} \\ \hline
\textbf{LSTM}  & \(2.446 \times 10^{-5}\) & \(4.060 \times 10^{-5}\) \\ \hline
\textbf{QLSTM} & \(1.041 \times 10^{-5}\) & \(1.347 \times 10^{-5}\) \\ \hline
\textbf{QGRU}  & \(5.386 \times 10^{-5}\) & \(6.342 \times 10^{-5}\) \\ \hline
\end{tabular}
\caption{Experiment 3: Final MAE and RMSE values for Methods, Variable \( v \).}
\label{tab:results_g2}
\end{table}

QLSTM consistently achieves the lowest MAE and RMSE across both datasets, outperforming traditional LSTM and QGRU. 

\begin{figure}[H]
    \centering
    \includegraphics[width=0.45\textwidth]{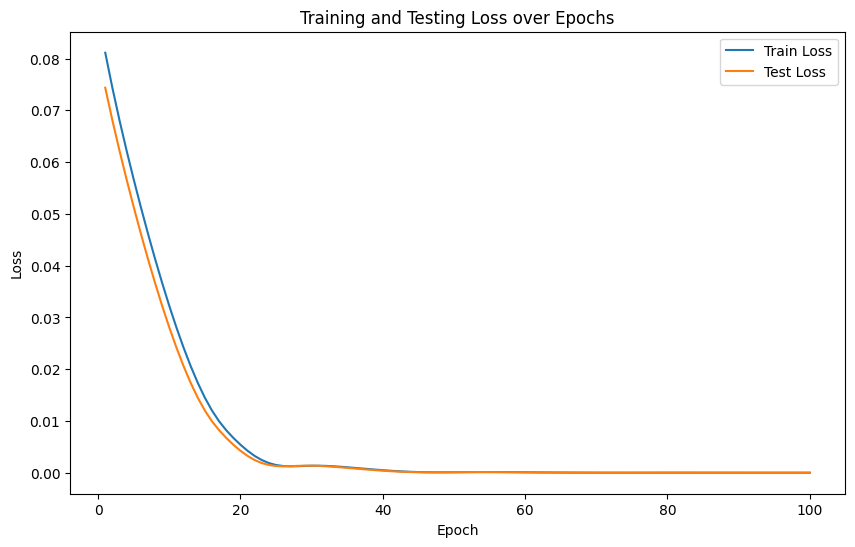}
    \caption{Training and test loss over epochs for LSTM.}
    \label{fig:log_loss_lstm_gs}
\end{figure}

\begin{figure}[H]
    \centering
    \includegraphics[width=0.45\textwidth]{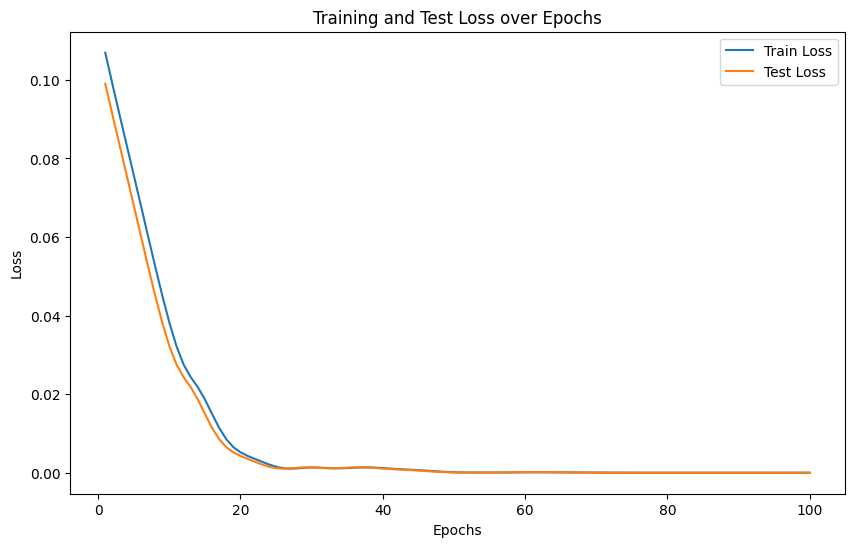}
    \caption{Training and test loss over epochs for QLSTM.}
    \label{fig:log_loss_qlstm_gs}
\end{figure}

\begin{figure}[H]
    \centering
    \includegraphics[width=0.45\textwidth]{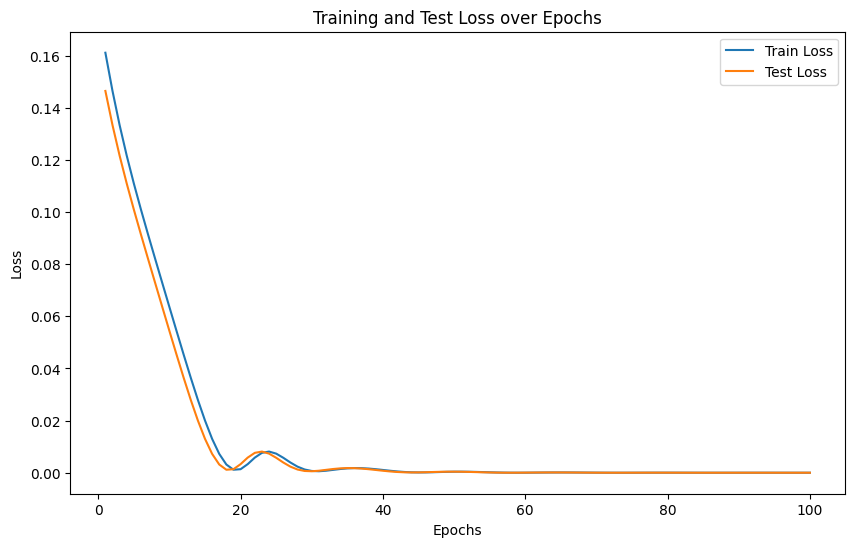}
    \caption{Training and test loss over epochs for QGRU.}
    \label{fig:log_loss_qgru_gs}
\end{figure}

\begin{figure}[H]
    \centering
    \begin{minipage}[b]{0.2\textwidth}
        \centering
        \includegraphics[width=\textwidth]{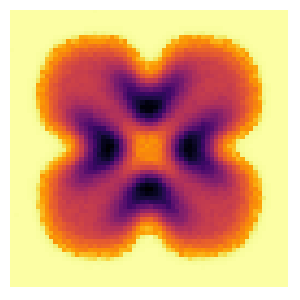}

    \end{minipage}
    \hfill
    \begin{minipage}[b]{0.2\textwidth}
        \centering
        \includegraphics[width=\textwidth]{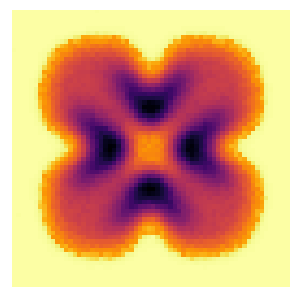}

    \end{minipage}
    \caption{Actual Data vs LSTM Heatmaps for variable u.}
    
\end{figure}

\begin{figure}[H]
    \centering
    \begin{minipage}[b]{0.2\textwidth}
        \centering
        \includegraphics[width=\textwidth]{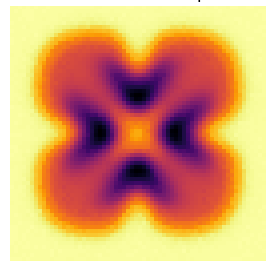}

    \end{minipage}
    \hfill
    \begin{minipage}[b]{0.2\textwidth}
        \centering
        \includegraphics[width=\textwidth]{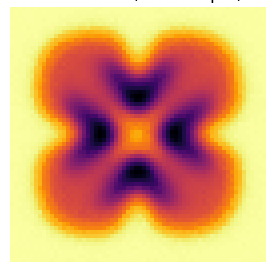}

    \end{minipage}
    \caption{Actual Data vs QLSTM Heatmaps for variable u.}
    
\end{figure}

\begin{figure}[H]
    \centering
    \begin{minipage}[b]{0.2\textwidth}
        \centering
        \includegraphics[width=\textwidth]{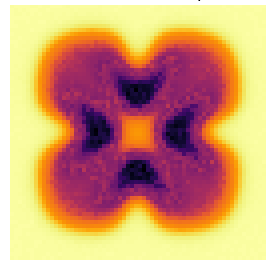}

    \end{minipage}
    \hfill
    \begin{minipage}[b]{0.2\textwidth}
        \centering
        \includegraphics[width=\textwidth]{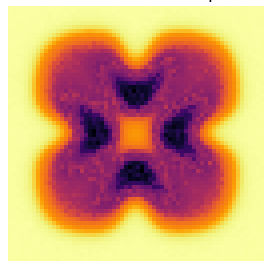}

    \end{minipage}
    \caption{Actual Data vs QGRU Heatmaps for variable u.}
   
\end{figure}

\begin{figure}[H]
    \centering
    \begin{minipage}[b]{0.2\textwidth}
        \centering
        \includegraphics[width=\textwidth]{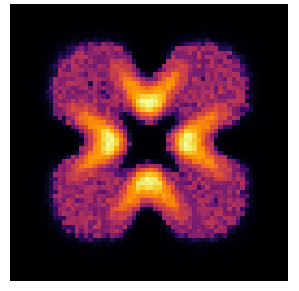}

    \end{minipage}
    \hfill
    \begin{minipage}[b]{0.2\textwidth}
        \centering
        \includegraphics[width=\textwidth]{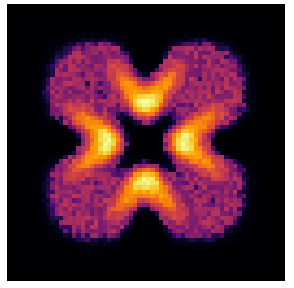}

    \end{minipage}
    \caption{Actual Data vs LSTM Heatmaps for variable v.}
    
\end{figure}

\begin{figure}[H]
    \centering
    \begin{minipage}[b]{0.2\textwidth}
        \centering
        \includegraphics[width=\textwidth]{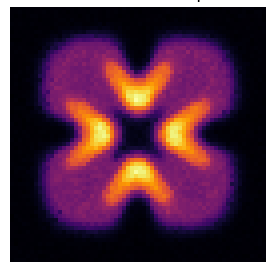}

    \end{minipage}
    \hfill
    \begin{minipage}[b]{0.2\textwidth}
        \centering
        \includegraphics[width=\textwidth]{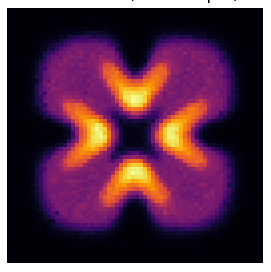}

    \end{minipage}
    \caption{Actual Data vs QLSTM Heatmaps for variable v.}
    
\end{figure}

\begin{figure}[H]
    \centering
    \begin{minipage}[b]{0.2\textwidth}
        \centering
        \includegraphics[width=\textwidth]{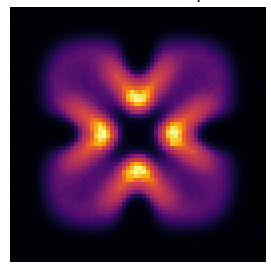}

    \end{minipage}
    \hfill
    \begin{minipage}[b]{0.2\textwidth}
        \centering
        \includegraphics[width=\textwidth]{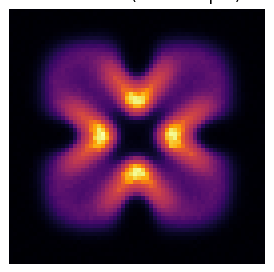}

    \end{minipage}
    \caption{Actual Data vs QGRU Heatmaps for variable v.}
    
\end{figure}

Figures 11-13 show the loss curves for all the models over epochs, smooth and consistent decreases in loss for each model are observed. Additionally, Figures 14-19 present example heatmaps of the spatial-temporal evolution of the exact and predicted 2D Gray-Scott solutions for all models, on variables \(u\) and \(v\). It is worth to mention, since the dataset was split randomly, the testing ground truth values may differ between models. To facilitate direct comparisons, separate heatmaps of the ground truth values are provided for each example alongside their respective predictions. Overall, all models demonstrate quite accurate predictions.

\subsection{Experiment 3: Hamilton-Jacobi-Bellman Equation}
For Experiment 3, we consider the 2D Hamilton-Jacobi-Bellman (HJB) equation \cite{bardicapuzzo1997}, a cornerstone in optimal control theory. The HJB equation used is:

\begin{align}
    \frac{\partial V}{\partial t} - \frac{1}{2} \left[ \left( \frac{\partial V}{\partial x} \right)^2 + \left( \frac{\partial V}{\partial y} \right)^2 \right] = 0,
\end{align}

where \( V(x, y, t) \) represents the value function describing the optimal cost-to-go. This equation is derived by minimizing the Hamiltonian:

\begin{align}
    \mathcal{H}(x, y, u^*, \nabla V) = -\frac{1}{2} \left[ \left( \frac{\partial V}{\partial x} \right)^2 + \left( \frac{\partial V}{\partial y} \right)^2 \right],
\end{align}

with the optimal control variables \( u_x^* \) and \( u_y^* \) expressed as:

\begin{align}
    u_x^* &= -\frac{\partial V}{\partial x}, \\
    u_y^* &= -\frac{\partial V}{\partial y}.
\end{align}

This equation captures the spatiotemporal evolution of the value function and serves as a benchmark for evaluating our proposed modeling approach.

We simulate the 2D HJB equation using a finite difference method. The computational domain is discretized into a grid of \( 50 \times 50 \) points over \( x \in [0, 1] \) and \( y \in [0, 1] \). The value function \( V(x, y, t) \) is evolved over \( 100 \) time steps with a time step \( dt = 0.0005 \).

The spatial gradients \( \frac{\partial V}{\partial x} \) and \( \frac{\partial V}{\partial y} \) are computed using central differences with appropriate boundary conditions. The Hamiltonian is calculated as:
\begin{equation}
    H = \frac{1}{2} \left[ \left( \frac{\partial V}{\partial x} \right)^2 + \left( \frac{\partial V}{\partial y} \right)^2 \right].
\end{equation}
The value function is updated:
\begin{equation}
    V^{n+1} = V^n - dt \cdot H,
\end{equation}
where \( V^n \) is the value function at time step \( n \). Boundary conditions are applied to ensure \( V = 0 \) at the domain boundaries.

The value function data generated by the simulation is preprocessed and normalized for training. An autoencoder is trained to reduce the dimensionality of the spatial data from \( 50 \times 50 = 2500 \) to a latent size of \( 16 \).

As seen in the Table \ref{tab:results_hjb}, QLSTM achieved the best results, with an MAE of \(1.365 \times 10^{-5}\) and an RMSE of \(1.980 \times 10^{-5}\). The QGRU also demonstrated good performance but slightly underperformed compared to QLSTM, with an MAE of \(3.183 \times 10^{-5}\) and an RMSE of \(4.548 \times 10^{-5}\).

\begin{table}[H]
\centering
\begin{tabular}{|c|c|c|}
\hline
\textbf{Models} & \textbf{MAE}       & \textbf{RMSE}      \\ \hline
\textbf{LSTM}  & \(1.741 \times 10^{-2}\) & \(2.585 \times 10^{-2}\) \\ \hline
\textbf{QLSTM} & \(1.365 \times 10^{-5}\) & \(1.980 \times 10^{-5}\) \\ \hline
\textbf{QGRU}  & \(3.183 \times 10^{-5}\) & \(4.548 \times 10^{-5}\) \\ \hline
\end{tabular}
\caption{Experiment 3: Final MAE and RMSE values for the models.}
\label{tab:results_hjb}
\end{table}

The following figures show the training and test loss over epochs for the models.
\begin{figure}[H]
    \centering
    \includegraphics[width=0.45\textwidth]{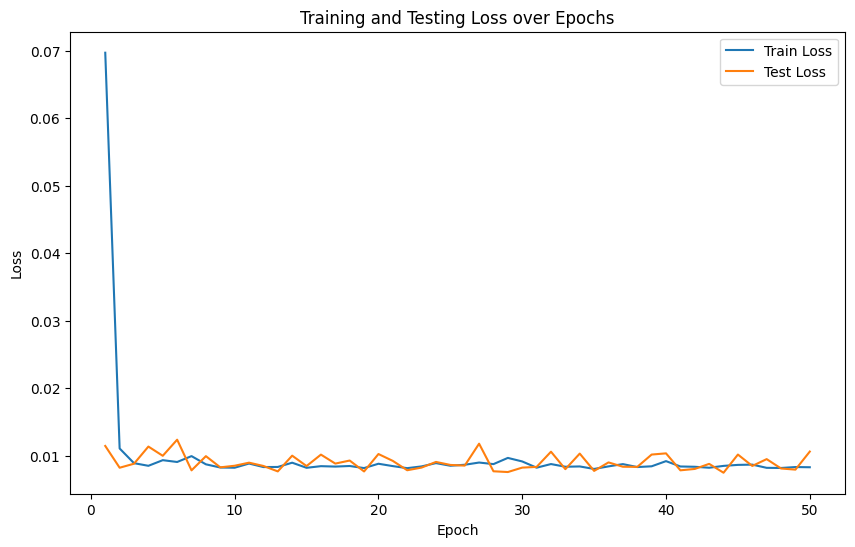}
    \caption{Training and test loss over epochs for LSTM.}
    \label{fig:log_loss_lstm_hjb}
\end{figure}

\begin{figure}[H]
    \centering
    \includegraphics[width=0.45\textwidth]{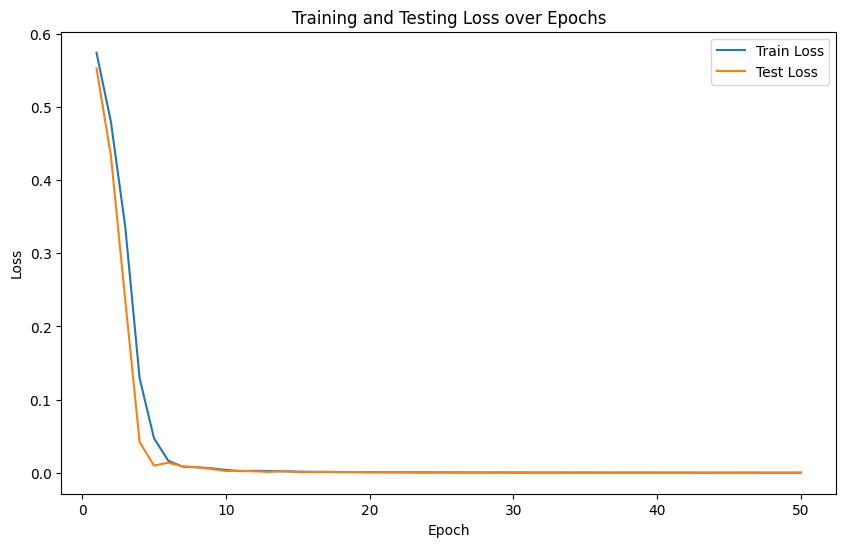}
    \caption{Training and test loss over epochs for QLSTM.}
    \label{fig:log_loss_qlstm_hjb}
\end{figure}

\begin{figure}[H]
    \centering
    \includegraphics[width=0.45\textwidth]{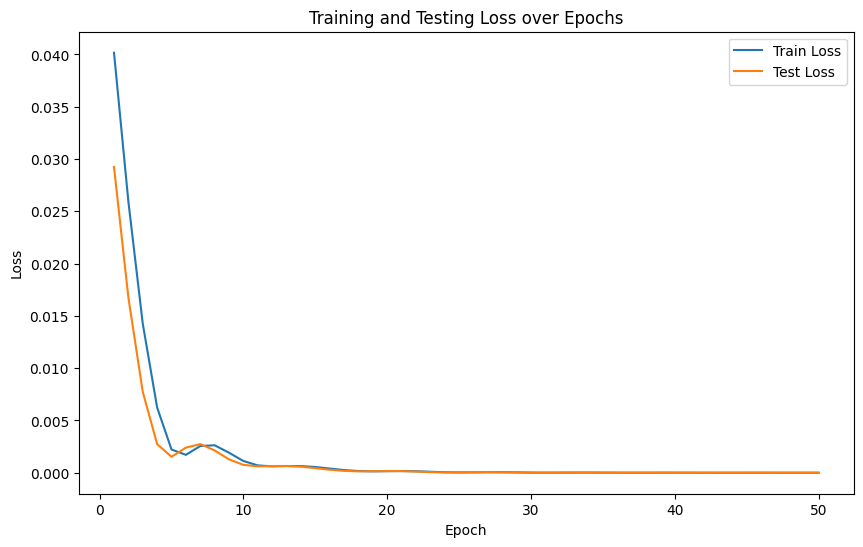}
    \caption{Training and test loss over epochs for QGRU.}
    \label{fig:log_loss_qgru_hjb}
\end{figure}

\vspace{1cm}

In the comparison of losses, the LSTM model exhibits unstable behavior, with noticeable oscillations and spikes throughout the epochs. In contrast, the quantum-based models consistently demonstrate smooth and steadily decreasing loss curves. To further evaluate the performance of these models, the actual and predicted value functions are compared at specific time steps. The following figures display 2D heatmaps and surface plots of the actual and predicted value functions for these selected time steps.

\begin{figure}[H]
    \centering
    \includegraphics[width=0.5\textwidth]{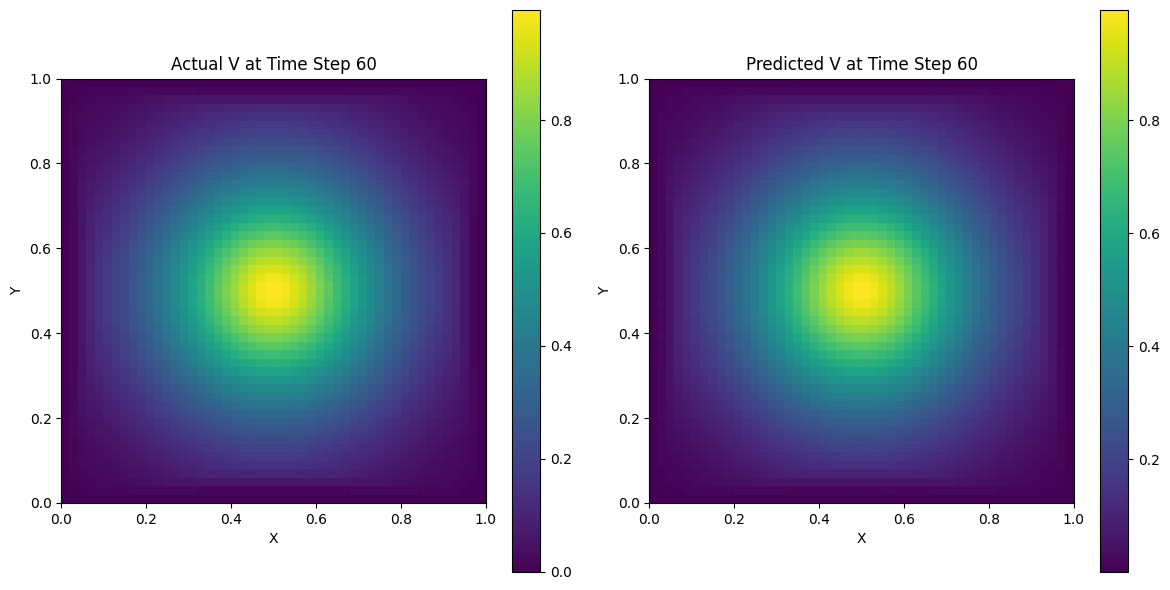}
    \caption{Heatmap: Actual vs LSTM.}
    \label{fig:heatmap_lstm_hjb}
\end{figure}

\begin{figure}[H]
    \centering
    \includegraphics[width=0.5\textwidth]{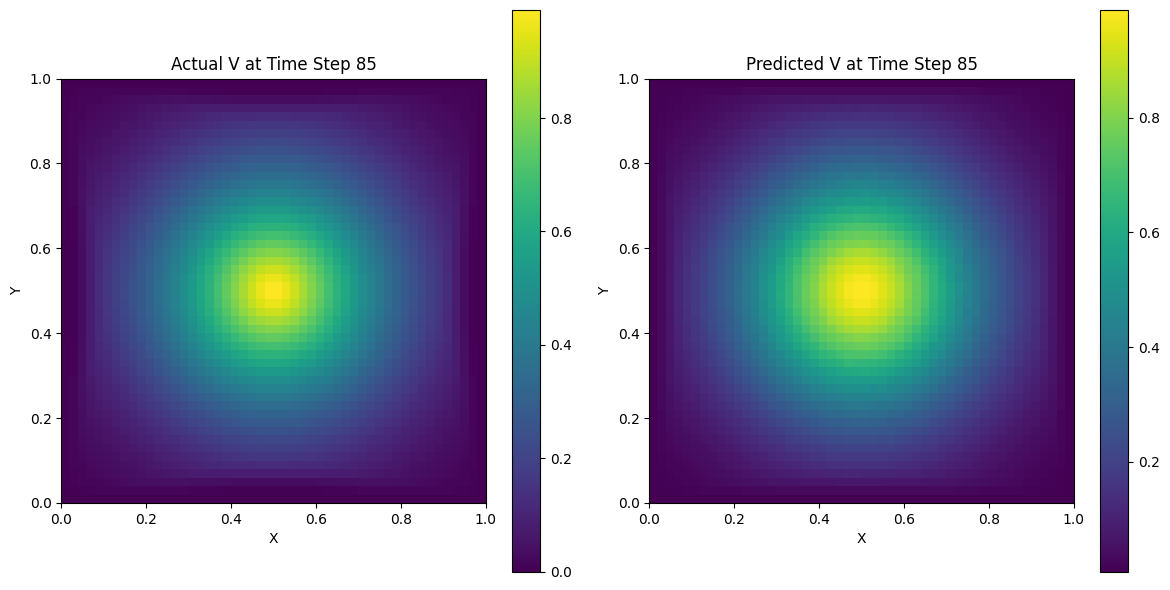}
    \caption{Heatmap: Actual vs QLSTM.}
    \label{fig:heatmap_qlstm_hjb}
\end{figure}

\begin{figure}[H]
    \centering
    \includegraphics[width=0.5\textwidth]{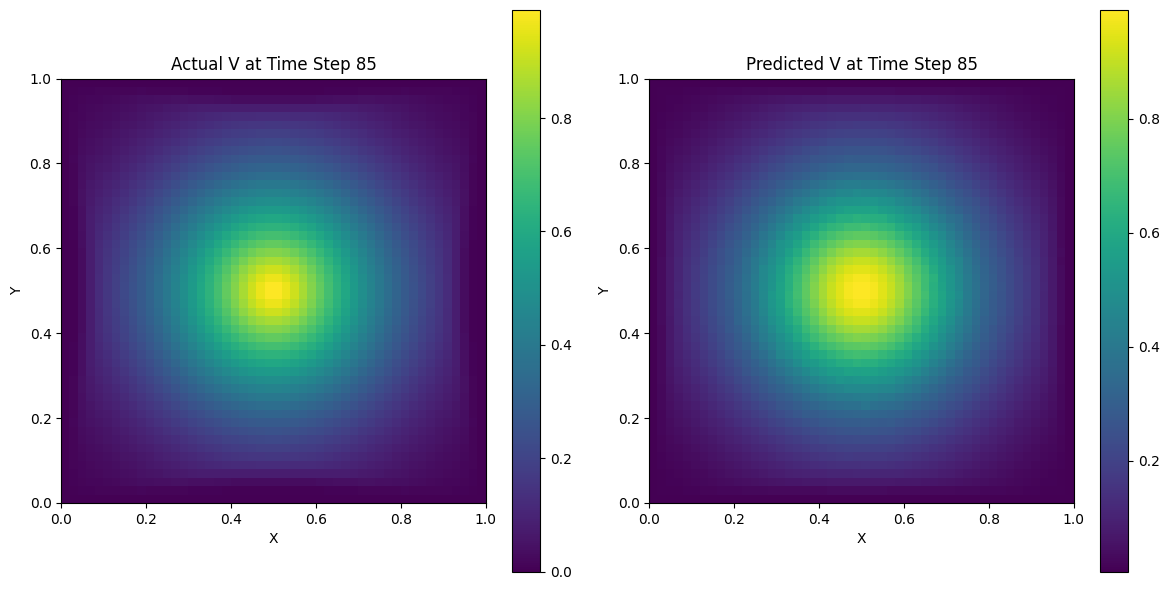}
    \caption{Heatmap: Actual vs QGRU.}
    \label{fig:heatmap_qgru_hjb}
\end{figure}

\begin{figure}[H]
    \centering
    \includegraphics[width=0.5\textwidth]{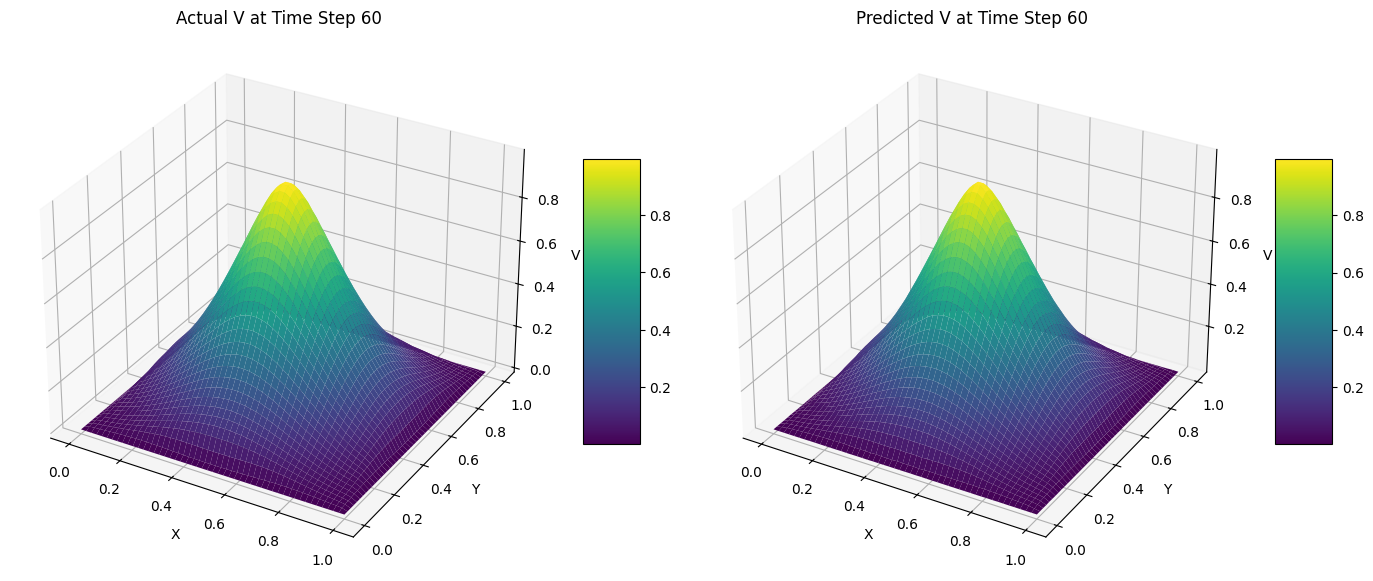}
    \caption{Surface plot: Actual vs LSTM.}
    \label{fig:surface_lstm_hjb}
\end{figure}

\begin{figure}[H]
    \centering
    \includegraphics[width=0.5\textwidth]{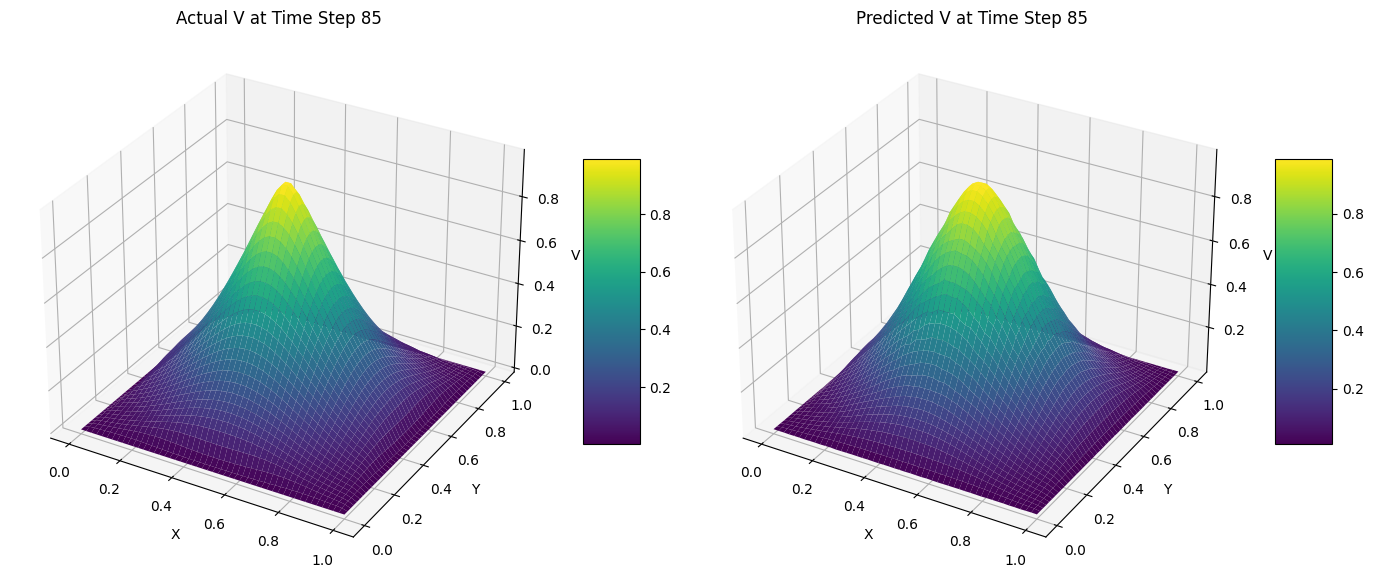}
    \caption{Surface plot: Actual vs QLSTM.}
    \label{fig:surface_qlstm_hjb}
\end{figure}

\begin{figure}[H]
    \centering
    \includegraphics[width=0.5\textwidth]{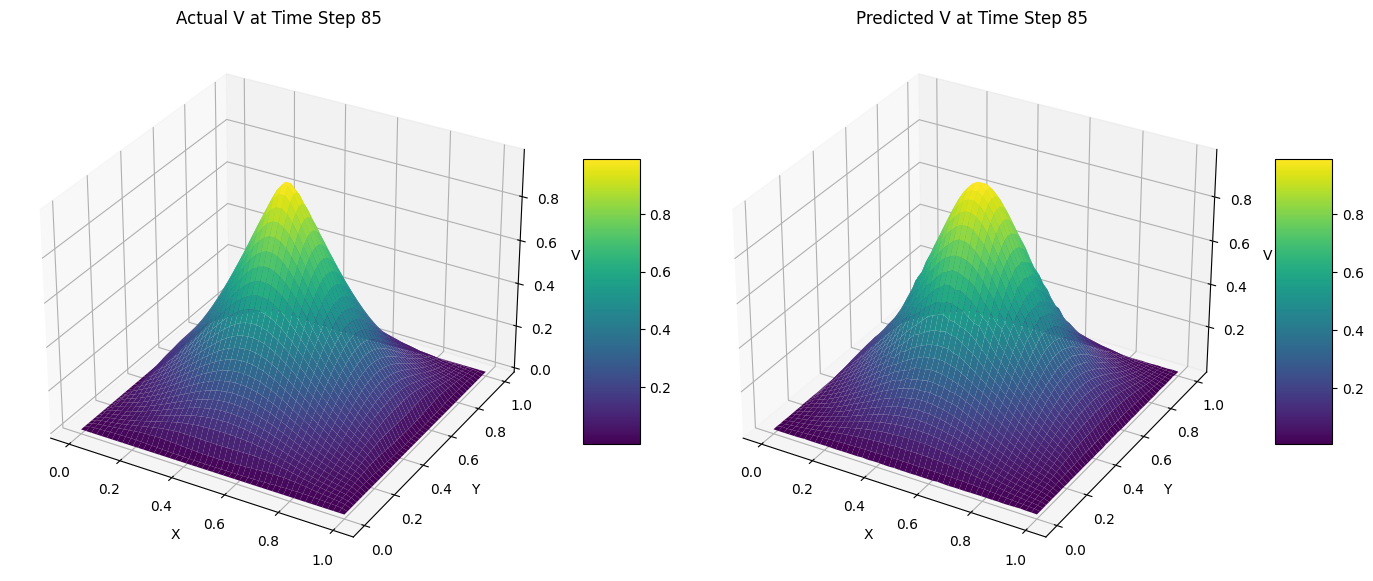}
    \caption{Surface plot: Actual vs QGRU.}
    \label{fig:surface_qgru_hjb}
\end{figure}

From the selected timestep plots, it is evident that the models successfully capture the overall trends, maybe some finer details are not fully represented, the performance is still promising. We will go one more step and test the higher dimension problem in the next experiment.

\subsection{Experiment 4: Michaelis–Menten Reaction-Diffusion System}

For Experiment 4, we consider the Michaelis–Menten reaction-diffusion system \cite{murray2002}, which models enzymatic reactions with diffusion. The governing equation is given by:

\begin{align}
    \frac{\partial u}{\partial t} = d \left( \frac{\partial^2 u}{\partial x^2} + \frac{\partial^2 u}{\partial y^2} + \frac{\partial^2 u}{\partial z^2} \right) - \frac{u}{1 + u},\\
    \quad 0 < x, y, z < 1, \; t > 0,
\end{align}

where \( u(x, y, z, t) \) represents the substrate concentration, \( d = 0.2 \) is the diffusion coefficient, and \( \frac{u}{1 + u} \) is the nonlinear reaction term. The computational domain is a unit cube, with homogeneous Dirichlet boundary conditions \( u(x, y, z, t) = 0 \) on the boundary, and the initial condition \( u(x, y, z, 0) = 1 \) in the interior.

The domain is discretized into a grid of \( 32 \times 32 \times 32 \), with spatial steps \( \Delta x = \Delta y = \Delta z = 1/31 \). The system is evolved over time using a finite difference method with a time step \( \Delta t = 10^{-4} \) up to \( t = 1.0 \), resulting in 10,000 time steps. Snapshots of the solution are recorded at 20 evenly spaced intervals for training and evaluation.

The resulting spatial data from each time step is flattened and normalized. An autoencoder is then employed to reduce the dimensionality of the data from \( 32^3 = 32,768 \) to a latent space of size 16. The autoencoder architecture consists of five layers in both the encoder and decoder, utilizing hyperbolic tangent activations for the hidden layers and a sigmoid activation for the output layer. The autoencoder is trained for 100 epochs with a batch size of 32.

\begin{table}[H]
\centering
\begin{tabular}{|c|c|c|}
\hline
\textbf{Models} & \textbf{MAE}       & \textbf{RMSE}      \\ \hline
\textbf{LSTM}  & \(5.812 \times 10^{-3}\) & \(9.661 \times 10^{-3}\) \\ \hline
\textbf{QLSTM} & \(4.041 \times 10^{-3}\) & \(5.061 \times 10^{-3}\) \\ \hline
\textbf{QGRU}  & \(3.151 \times 10^{-3}\) & \(3.418 \times 10^{-3}\) \\ \hline
\end{tabular}
\caption{Experiment 4: Final MAE and RMSE values for the models.}
\label{tab:resultsexp3}
\end{table}

For this experiment, all the models are trained for 200 epochs, the following figures illustrate the training and test losses for the three models over the epochs.

\begin{figure}[H]
    \centering
    \includegraphics[width=0.4\textwidth]{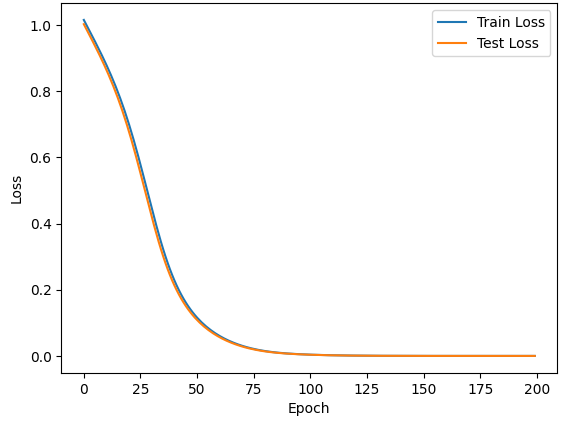}
    \caption{Training and test loss over epochs for LSTM.}
    \label{fig:log_loss_lstm_mm}
\end{figure}

\begin{figure}[H]
    \centering
    \includegraphics[width=0.4\textwidth]{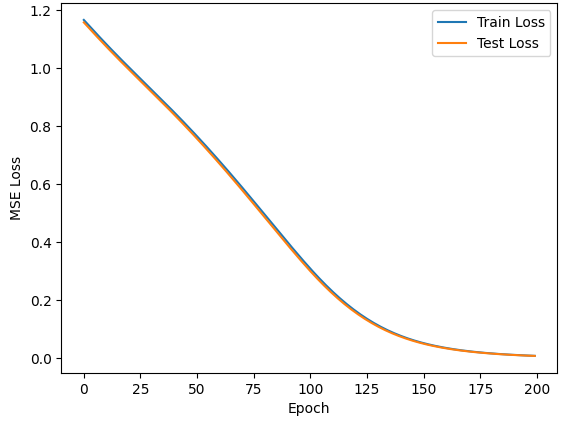}
    \caption{Training and test loss over epochs for QLSTM.}
    \label{fig:log_loss_qlstm_mm}
\end{figure}

\begin{figure}[H]
    \centering
    \includegraphics[width=0.4\textwidth]{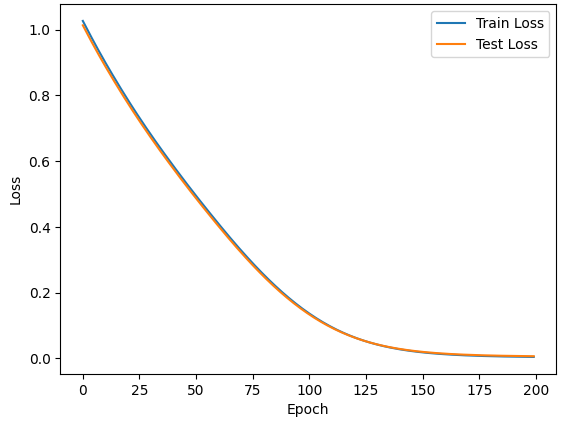}
    \caption{Training and test loss over epochs for QGRU.}
    \label{fig:log_loss_qgru_mm}
\end{figure}

After confirming that the models exhibit stable and consistently decreasing losses, we examine the actual predicted values. Specifically, we compare the reconstructed 3D concentration fields at the predicted slices with \(z = nz/2\). The following figures present the surface plots of the actual and predicted concentration fields.

\begin{figure}[H]
    \centering
    \includegraphics[width=0.4\textwidth]{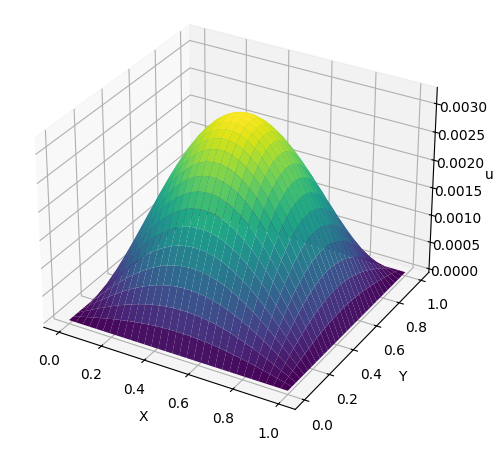}
    \caption{Predicted Slices Ground Truth.}
    \label{fig:actual_mm}
\end{figure}

\begin{figure}[H]
    \centering
    \includegraphics[width=0.4\textwidth]{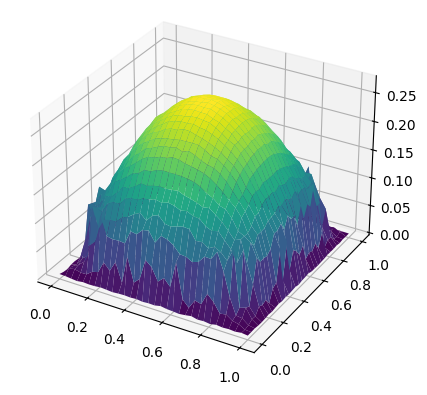}
    \caption{Predicted Slices LSTM.}
    \label{fig:lstm_mm}
\end{figure}

\begin{figure}[H]
    \centering
    \includegraphics[width=0.4\textwidth]{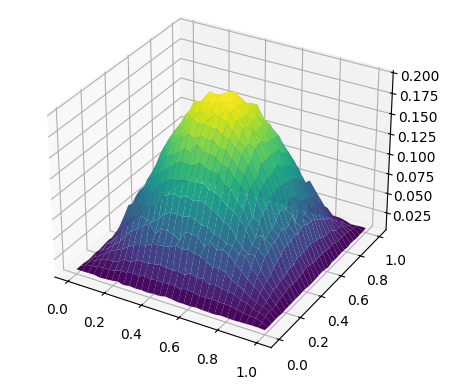}
    \caption{Predicted Slices QLSTM.}
    \label{fig:qlstm_mm}
\end{figure}

\begin{figure}[H]
    \centering
    \includegraphics[width=0.4\textwidth]{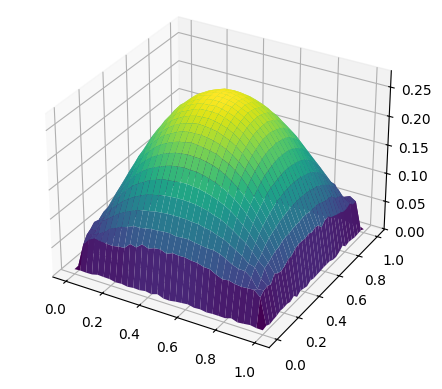}
    \caption{Predicted Slices QGRU.}
    \label{fig:qgru_mm}
\end{figure}

From the surface plots, it is clear that the models face more challenges compared to the previous 2-D problems. The LSTM model produces predictions with noticeable artifacts, particularly at the edges, indicating difficulty in capturing the overall structure. Similarly, QGRU shows artifacts, but its predictions are significantly smoother than those of the LSTM, especially at the peak of the smaller hill. Among all the models, QLSTM demonstrates the best overall shape, although it still requires further refinement to capture finer details accurately.

It is important to note that since the higher-dimensional data are encoded using the autoencoder before performing time series predictions in the latent space, the reconstructed results are highly dependent on the performance of the encoder-decoder. The effectiveness of the predictions, therefore, relies on the autoencoder's ability to preserve key features of the original data during encoding and reconstruction for this algorithm.
\section{Conclusion}
The encoder-decoder Quantum Recurrent Neural Networks algorithms, specifically QLSTM and QGRU, provide a promising framework for solving nonlinear time-dependent PDEs by integrating Variational Quantum Circuits into recurrent architectures. Results from four numerical experiments demonstrate that QRNNs achieve competitive accuracy, with QLSTM consistently outperforming classical LSTM in stability and predictive precision. 

As quantum hardware advances, QRNNs have the potential to become a useful tool in computational science, offering a novel approach for efficiently solving high-dimensional PDEs while improving accuracy, and stability.

\end{document}